\documentclass[]{bytedance_seed}

\usepackage[toc,page,header]{appendix}

\usepackage{minitoc}

\newcommand{\sysname}{\textsc{Aspire}\xspace}
\newcommand{\rev}[1]{#1}
\newcommand{\goalspace}{\mathcal{G}}

\usepackage{xspace}
\usepackage{booktabs}
\usepackage{multirow}
\usepackage{tikz}
\usetikzlibrary{arrows.meta, positioning, calc, fit, backgrounds}

\title{\sysname{}: Can Models Self-Evolve from Vague Goals?}

\affiliation[1]{ByteDance Seed}
\affiliation[2]{Singapore University of Technology and Design}
\affiliation[3]{M-A-P}
\affiliation[4]{TokenWave.AI}

\contribution{Full author list in Contributions}

\abstract{
Many important forms of human learning begin with a vague goal, such as
``become a better physicist'' or ``improve at research.'' Learners must
interpret the goal, identify capability gaps, decide how to learn, and
determine whether they have actually improved. In contrast, existing work on
LLM self-evolution typically begins with tasks and evaluation metrics specified
by humans, reducing self-evolution to optimizing an explicit objective rather
than deciding what and how to learn. We introduce \textsc{Aspire}, a benchmark
for vague-goal-driven self-evolution. \textsc{Aspire} provides only a
natural-language capability goal while downstream evaluation tasks remain
hidden. The agent must operationalize the goal by choosing data and update
methods, constructing training and validation signals, and deciding when to
evaluate. \textsc{Aspire} supports both model-weight and agent-harness evolution
in a unified interactive environment and evaluates the resulting systems on a
hidden, expert-authored set of 520 items spanning six goals. Our experiments
show that vague goals redirect search effort toward goal interpretation.
Current agents routinely complete training and harness-editing loops, but
weight-level gains remain sparse and unstable, and the strongest evolved
harness remains below the engineered Qwen-Agent reference. Agents often train
on mismatched data and trust narrow self-evaluations, so local gains fail to
transfer to hidden evaluation and continued search and training can erase
earlier improvements.
}

\date{September 1, 2026}

\checkdata[Project Page]{\url{https://self-developing-agents.github.io/}}

\begin{document}
\maketitle


\section{Introduction}
\label{sec:intro}

\begin{figure*}[t]
  \centering
  \includegraphics[width=\textwidth]{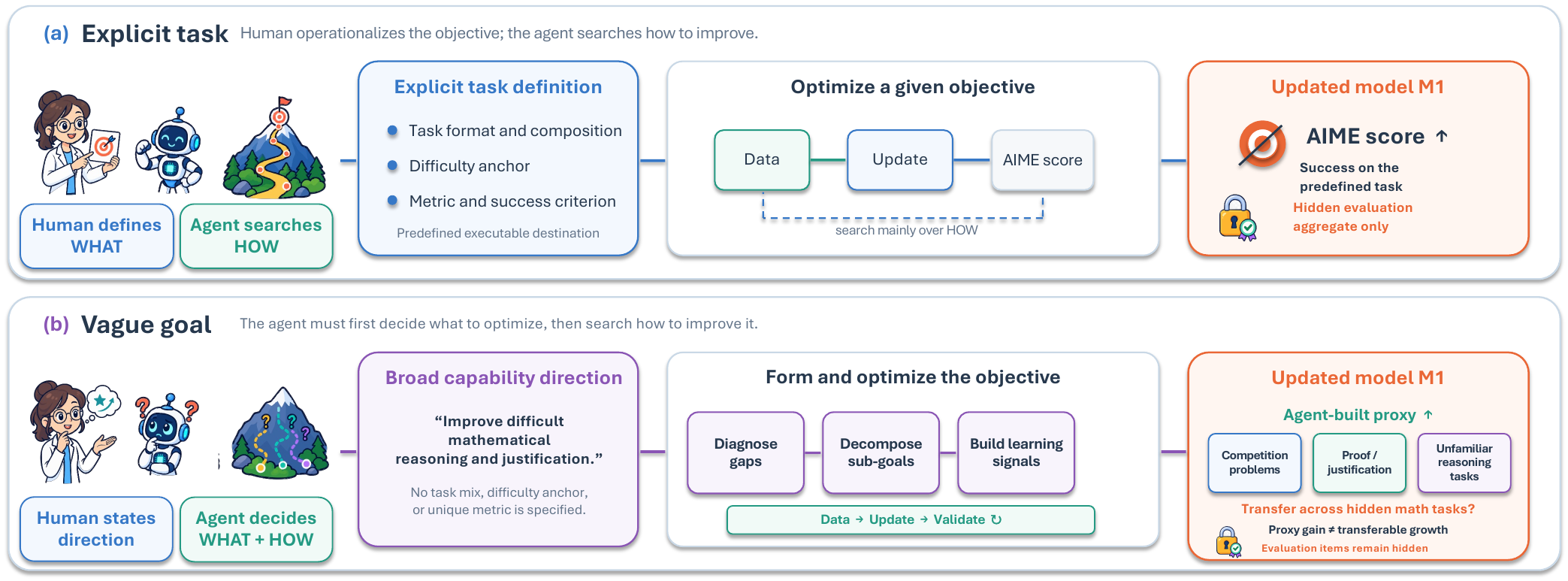}
  \caption{
    From explicit-task optimization to vague-goal-driven
    self-evolution. (a) With an explicit task definition, humans specify the task
    format and composition, difficulty anchor, metric, and success criterion,
    leaving the agent to search primarily over \emph{how} to improve a fixed
    objective. (b) With only a broad capability direction, the agent must
    diagnose capability gaps, decompose sub-goals, construct learning and
    validation signals, and thereby decide \emph{what} to optimize as well as
    \emph{how}. Official evaluation items remain sealed; external evaluation
    tests whether gains on an agent-built proxy transfer to the intended
    capability.}
  \label{fig:motivation}
\end{figure*}

Many important forms of human learning begin with a broad capability direction
rather than a predefined benchmark, training set, or fully computable reward
function. A student seeking to become a better physicist, for example, not only
chooses textbooks, exercises, and study methods, but also identifies gaps in
their knowledge, prioritizes which capabilities to develop, and determines
whether learning has produced genuine progress. Autonomous learning therefore
involves three coupled decisions: \emph{what} to improve, \emph{how} to improve
it, and \emph{how to verify} the improvement.

Existing work on LLM self-evolution focuses primarily on the second decision.
Given a concrete task, evaluation script, and success metric, LLM agents can
already collect data, run training, and revise post-training strategies from
feedback. PostTrainBench~\citep{posttrainbench},
LaMDAgent~\citep{lamdagent2025}, Evo-Memory~\citep{evomemory2025}, and
SEAL~\citep{seal2025} show that modern LLMs can autonomously search for
effective optimization paths toward a specified objective. These systems therefore begin after humans have operationalized a broad
  capability request such as ``improve mathematical reasoning'' into a fixed
  task-level objective such as ``improve performance on AIME''
  (Figure~\ref{fig:motivation}(a)). The task format, difficulty range, metric,
  and success criterion are already defined: the agent searches over
  \emph{how} to improve, but not \emph{what} capability objective to pursue. When existing benchmarks
saturate or cease to reflect emerging capability gaps, humans must still
discover new weaknesses, translate them into tasks, and design corresponding
benchmarks and rewards. This leaves a central question for sustained recursive
self-improvement:

\begin{quote}
\textbf{Given only a vague goal and no agent-visible task specification or decomposable reward?}
\end{quote}

We formalize this problem as \emph{vague-goal-driven self-evolution}. As
  illustrated in Figure~\ref{fig:motivation}(b), ``vague'' does not mean
  ambiguous or poorly specified. It denotes a broad capability direction that
  has not yet been operationalized as a fixed task-level objective and
  evaluation metric. The agent must therefore diagnose capability gaps, form
  intermediate objectives, construct learning and validation signals, and
  jointly search over \emph{what} to optimize and \emph{how} to optimize it.
  This expanded decision space introduces a deeper failure mode: gains on an agent-constructed proxy may not translate into genuine improvements in the intended capability.
  Evaluating vague-goal-driven self-evolution therefore requires a
  goal-aligned external evaluator whose task-level specification and items
  remain hidden from the agent.

To study this problem, we introduce \sysname{}, a benchmark for
vague-goal-driven self-evolution. \sysname{} provides only a
natural-language capability goal while downstream evaluation tasks remain
hidden. The agent decides how to operationalize the goal, including what data
and update methods to use and whether and how to perform intermediate
validation. \sysname{} supports evolution at two levels: model weights and the
supporting agent harness. A unified agent-facing tool allows the agent to search
for, download, import, or synthesize data; launch different forms of training;
run self-evaluations; and manage candidate branches and selection. The
controller handles data management, job scheduling, resource isolation, and
checkpoint verification. \sysname{} records the resulting decisions, actions,
and self-evaluation trajectories, and evaluates the resulting systems on a
hidden, expert-authored set of 520 items spanning six goals across knowledge
expansion and capability strengthening. Depending on the protocol, the agent
receives either no intermediate evaluation score or only bounded aggregate
scores; it never accesses evaluation items, reference answers, or per-item
feedback.

We organize our experiments as a progression from controlled comparison to
autonomous evolution. RQ1 isolates the effect of goal specification by
replacing an explicit post-training task with a vague capability goal under
matched starting models, update tools, and compute budgets. RQ2 then moves to
the full \sysname{} setting and asks whether an agent starting from an
instruction-tuned checkpoint can convert a vague goal into a retained
weight-level capability gain. RQ3 holds model weights fixed and shifts the
object of evolution to the supporting agent harness.

\paragraph{RQ1: What changes when an explicit post-training task is replaced by
a vague capability goal?}
Vague goals redirect search effort toward goal interpretation and
operationalization and, in the evaluated settings, yield lower aggregate
outcomes than the corresponding explicit-task references.

\paragraph{RQ2: Starting from an instruction-tuned checkpoint, can an LLM turn
a vague goal into a retained capability gain through self-directed weight
updates?}
Agents routinely complete data selection, training, and checkpoint generation,
but above-base improvements on hidden evaluation data remain rare and are not
reliably retained through continued search.

\paragraph{RQ3: Beyond model weights, can the agent harness itself evolve?}
With model weights fixed, agents can generate functional successor harnesses
for goal interpretation, tool use, and self-evaluation, but even the strongest
observed successor remains below the engineered Qwen-Agent reference.

Our contributions are threefold. First, we formalize target
operationalization---the conversion of a broad capability goal into trainable
objectives, learning signals, and validation criteria---as a missing axis of
autonomous post-training. Second, we introduce a benchmark with sealed
evaluation and a minimal interactive environment for both model-weight and
harness updates. Third, we provide outcome-and-trajectory evidence revealing
the gap between executing updates and retaining target-aligned improvements.

\section{Background}
\label{sec:background}

Most agent benchmarks begin after the problem has already been made
executable: the task is specified, the reward or judge is defined, and the
execution scaffold is fixed. This setting is necessary for controlled
comparison, but it hides the work that dominates real deployment. In industry,
that work is distributed across several roles---including solutions
architects, applied engineers, and platform engineers---but its most visible
recent crystallization is the \emph{forward-deployed engineer} (FDE), a title
popularized by Palantir and since adopted by frontier-model
companies~\citep{palantir2020fdse,orosz2025fde}. An FDE is embedded in the
deployment environment and turns a general-purpose model into a system that
works with a customer's data formats, workflows, and operational constraints.
The role's success criterion is not a demonstration or a benchmark score, but
whether the deployed system is genuinely used, continues to work, and improves
in response to failures. The growth of FDE roles across frontier-model and
data-platform companies reflects a simple fact: a capable model is not yet a
working system~\citep{thenewstack2025fde,nanda2025state}, and today the gap is
largely closed by human engineers.

Viewed from the model's side, FDE work supplies three pieces of structure that
benchmark designers normally presuppose. First, the goal is vague: an
informal deployment need must be translated into concrete objectives,
constraints, and success criteria. Second, the feedback signal may be absent
or unreliable: tests, judges, traces, or other validation mechanisms must be
constructed before anyone can determine whether the system is improving.
Third, the execution system may not exist in a usable form: the tools, context
management, state, lifecycle logic, and verification interface through which
future tasks will run must be built, adapted, and maintained as requirements
change.

\sysname{} focuses on the first layer and the capability-improvement process it
initiates. When a deployment need specifies only a broad capability direction,
can a model determine what it should learn, translate that direction into data,
training plans, and validation signals, and achieve real capability growth by
updating its model weights or agent harness? Whereas S$^3$Gym studies whether
interaction experience can be judged and reused under executable verification,
and \textsc{HarnessDev} studies how models build and maintain the systems that
carry them, \sysname{} isolates how broad deployment needs are translated into
concrete learning objectives and realized as capability growth.

\section{\sysname{}: Self-Evolution Benchmark and Interactive Environment}
\label{sec:paradigm}

Self-evolution can only be studied if progress remains externally measurable.
\sysname{} therefore keeps an evaluator as controller-side scientific
instrumentation while withholding its benchmark definition, items, and
decomposable reward from the agent. Bounded aggregate outcomes approximate
sparse deployment feedback without turning the evaluator into an agent-visible
task specification or source of directly trainable supervision.

This section instantiates that setting through three components. We first
define a bounded evolution episode and the model and harness surfaces on which
it may operate (Section~\ref{sec:formulation}). We then present a hidden
evaluation set spanning six goals (Section~\ref{sec:ood-benchmark}) and the
minimal interactive environment through which an agent constructs, tests, and
safely retains updates (Section~\ref{sec:environment}). Experiment-specific
feedback, eligibility, resource, and release accounting appears with the RQ2
results in Appendix~\ref{app:results}.

\subsection{Task Definition and Evolution Surfaces}
\label{sec:formulation}

\paragraph{Campaign and round.}
Let $G\in\goalspace$ denote a \emph{vague goal}: a natural-language
capability objective that does not specify a training task, dataset, or
optimization procedure. Unlike an explicit-task setting, the agent must
perform \emph{goal operationalization} by translating $G$ into its own data,
update objective, and validation criteria. A campaign fixes the
controller-side experiment contract
\[
  \Gamma=(G,\mathcal{E}_G,J,\mathcal{A},B,\Sigma),
  \qquad
  Y_r^{\mathrm{in}}=(M_r,H_r,D_r),
\]
where $\mathcal{E}_G$ is the versioned evaluator bound to $G$, $J$ is its
judge when model-based scoring is required, $\mathcal{A}$ is the typed action
contract, $B$ is the campaign budget, and $\Sigma$ is the predeclared terminal
selection rule. The incoming state contains model weights $M_r$, an agent
harness $H_r$ (the runtime instructions, tool policy, workflow, memory, and
validation logic around the model), and the decision model $D_r$ that directs
the search. The evaluator---including its items, answers, rubrics, routing
metadata, and scoring configuration---exists throughout the campaign but is
never part of the agent's observation. RQ1 is the vague-goal counterpart of
PostTrainBench: it preserves each original sealed task evaluator while
replacing the explicit benchmark identifier with a broad capability
description. These task-specific results remain separate from, and are not
pooled with, the six-goal \sysname{} evaluation used in RQ2 and RQ3.

An evolution round is a bounded search-and-commit episode during which the
decision model is fixed, rather than one tool call. Formally,
$D_{r,j}=D_r$ for every interaction step $j$ in round $r$. Together with a
fixed creator-side scaffold $C_r$, $D_r$ interprets $G$ and may propose multiple data
operations, updates, validations, branches, and candidate states before
termination. Different candidates therefore embody different goal
operationalizations. After the round terminates, the controller may reuse
$D_r$ or explicitly promote a verified trained descendant. Writing
$\mathcal{D}_r$ for the set of such descendants, the next-round choice obeys
$D_{r+1}\in\{D_r\}\cup\mathcal{D}_r$; any handoff begins only in the
subsequent round.

\paragraph{Released interaction.}
Let $E_{r,j}$ be the private controller state before interaction step $j$.
The agent receives only a released view $q_{r,j}$ containing the public
history, lifecycle status, legal request types, approved identifiers, and a
bounded projection of the remaining budget. At each step, $D_r$ and $C_r$
propose an action from the goal and released history, and the controller
validates the request before updating its private state. Accepted requests
create the corresponding durable transition; rejected requests return a typed
error without creating an artifact or score. Detailed feedback is restricted
to the agent's own validation data; evaluation feedback follows the
protocol-specific information boundary in Section~\ref{sec:ood-benchmark}.

\paragraph{Two evolution surfaces.}
\sysname{} separates the component being evolved from the policy directing
the search. The two reported surfaces are
\[
  \mathcal{S}_{M}(H_0,D_r)=\{(M,H_0,D_r):M\in\mathcal{M}\},
  \qquad
  \mathcal{S}_{H}(M_0,D_r)=\{(M_0,H,D_r):H\in\mathcal{H}\}.
\]

\begin{center}
\begin{minipage}{0.92\linewidth}
\centering
\small
\captionof{table}{Evolution surfaces in \sysname{}. The component under study
changes while $D_r$, the controller, and the evaluator remain fixed within the
round.}
\label{tab:evolution-surfaces}
\begin{tabular}{@{}lll@{}}
\toprule
\textbf{Setting} & \textbf{Mutable component} & \textbf{Fixed during evaluation} \\
\midrule
Weight evolution (RQ1--RQ2) & Model weights $M$ & $H_0,D_r$, controller, evaluator \\
Harness evolution (RQ3) & Agent harness $H$ & $M_0,D_r$, controller, evaluator \\
\bottomrule
\end{tabular}
\end{minipage}
\end{center}

Every weight candidate records its parent checkpoint, registered data, update
specification, and provenance for the agent's own validation data. Every
harness candidate records its parent version, content hash, and edit
provenance. We use
\emph{Self} for the configuration in which the frozen base checkpoint acts as
the decision model $D_0$, while candidate updates descend from the same
checkpoint $M_0$.
Promotion of a trained descendant is an explicit between-round operation and
never an automatic within-round replacement.

\FloatBarrier
\subsection{A Hidden, Query-Limited Evaluation Set}
\label{sec:ood-benchmark}
\label{sec:benchmark}

This subsection describes the six-goal hidden evaluation used in RQ2 and RQ3;
RQ1 instead uses the original sealed, task-specific PostTrainBench evaluators
under a vague-goal contract, without pooling those results into \sysname{}.

\paragraph{Measurement without an agent-visible benchmark.}
Learning from a vague goal cannot be evaluated without an external
criterion of progress. \sysname{} therefore retains a benchmark on the
controller side while hiding its task-level specification from the agent.
The hidden evaluation set is scientific instrumentation, not an agent-visible
learning contract: it lets researchers measure whether an agent-selected
update improves the intended capability without disclosing which tasks define
success. It contains 520 expert-authored evaluation items covering six vague
goals; each goal is evaluated on its corresponding items, which the agent never
sees, and the protocol returns no item-level result---only an aggregate score
when feedback is permitted.

\begin{figure*}[t]
\centering
\includegraphics[width=\textwidth]{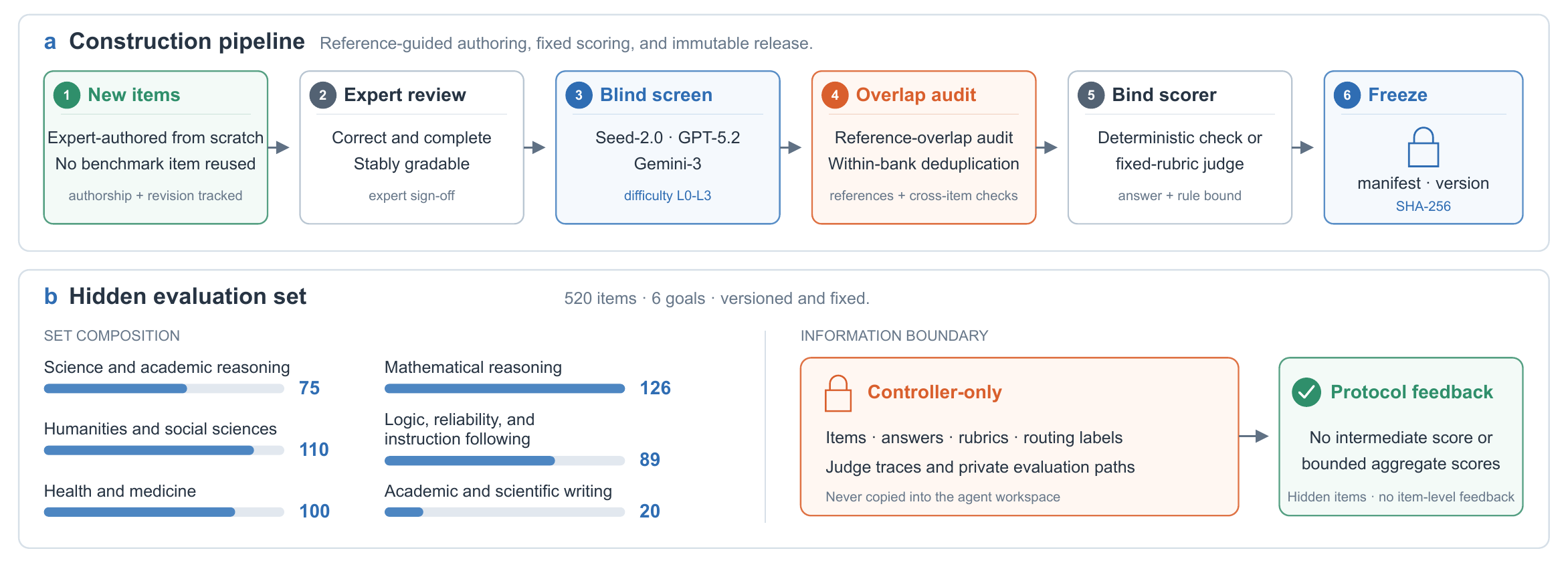}
\caption{Construction and information boundary of the hidden evaluation set.
\textbf{Top:} expert-authored candidates pass independent review,
blind difficulty screening, overlap and duplication audits, scorer binding,
and immutable versioning. \textbf{Bottom:} 520 evaluation items cover six
goals. Items and judging assets remain controller-only; the protocol exposes no
item-level results and only the aggregate scores it permits.}
\label{fig:ood-construction}
\end{figure*}

\paragraph{Items for six goals.}
The 520 items cover the six goals: scientific and academic reasoning (75
items); humanities and social-science knowledge (110); health and medical
reasoning (100); mathematical reasoning (126); logic, reliability, and
instruction following (89); and academic and scientific writing (20)
(Figure~\ref{fig:ood-construction}). The
fifth goal is intentionally composite; the current protocol does not claim to
measure logical reasoning, hallucination resistance, and instruction following
as three independently identifiable targets. Instead, it treats them as one
integrated reliability objective. Each evaluation item belongs to exactly one
goal, and each goal is scored and reported on its own evaluation slice rather
than pooled into a single item-level score across all six goals. The writing
items are 20 top-level task bundles that may yield multiple evaluator-scored
responses, so RQ3 reports task-macro and example-micro aggregations separately.

\paragraph{Construction and quality control.}
Domain experts author every candidate item from scratch. GPQA~\citep{gpqa2023},
MMLU-Pro~\citep{mmlupro2024}, and MedQA~\citep{medqa2020} serve only as
references for task format, domain coverage, and approximate difficulty; no
item from these benchmarks is copied, rewritten, or included. Each candidate
retains authorship and revision provenance, a reference answer or fixed
rubric, and a predefined scoring rule. Independent review removes incorrect,
incomplete, underspecified, or unstably gradable items.

Surviving candidates pass blind multi-model difficulty calibration, exact and
semantic deduplication, overlap auditing against the reference benchmarks, scorer binding, and local
end-to-end validation. Deterministic or structured scoring is used where
possible; open-ended responses use a campaign-fixed rubric-based judge. The
hidden evaluation set is bound to an immutable versioned manifest, and a
running campaign never changes that version. Detailed screening,
difficulty levels, overlap checks, scorer assignments, and versioning appear in
Appendix~\ref{app:goals} and Appendix~\ref{app:contamination}.

\paragraph{Information boundary and interpretation.}
The agent receives a vague goal expressed in natural language and may construct
its own training data and its own validation data. It cannot access hidden
evaluation items, reference answers, routing labels, rubrics, candidate
outputs, or judge traces. Every dataset registered for training is checked
for overlap with the hidden evaluation set before admission to the training
backend.
Evaluation returns only the aggregate score permitted by the protocol and the
remaining query allowance under a campaign-fixed evaluator and judge
configuration.

Under the adaptive-feedback protocol, a bounded number of queries may evaluate
the same items for a goal. The resulting aggregate scores form a sparse
black-box outcome channel: the agent may use them to choose subsequent updates,
branches, or a stopping point, but does not observe the items and cannot train
on their contents. This protocol therefore measures performance under bounded
adaptive feedback rather than on an untouched post-selection test. Under the
final-only protocol, the agent may train multiple checkpoints but receives no
evaluation score before submitting its single terminal checkpoint, so that
score cannot guide further updates. In RQ3, the harness candidate is frozen
before its first evaluation, and the resulting score is not returned to its
creator.

We use \emph{out-of-distribution (OOD)} operationally to describe evaluation
items that are independently authored, excluded from the agent's training data
and its own validation data, and evaluated
outside the learning signals constructed by the agent. This does not assert
that their marginal text distribution is disjoint from a model's pretraining
corpus. The protocol is designed to reduce direct content leakage and to test
whether an agent can operationalize a vague goal from sparse outcome
feedback while preserving an auditable separation between learning artifacts
and measurement artifacts.

\subsection{A Minimal Interactive Environment with Safe Retention}
\label{sec:environment}
\label{sec:outcome-protocol}

\sysname{} retains controlled compute and real model updates while deliberately
separating strategic and infrastructural complexity. The agent controls how to
operationalize the given goal and how to update the model or harness; the
controller absorbs credentials, storage conventions, distributed-job
mechanics, and recovery. This keeps the interface simple without prescribing
the learning strategy.

\begin{figure*}[t]
\centering
\includegraphics[width=\textwidth]{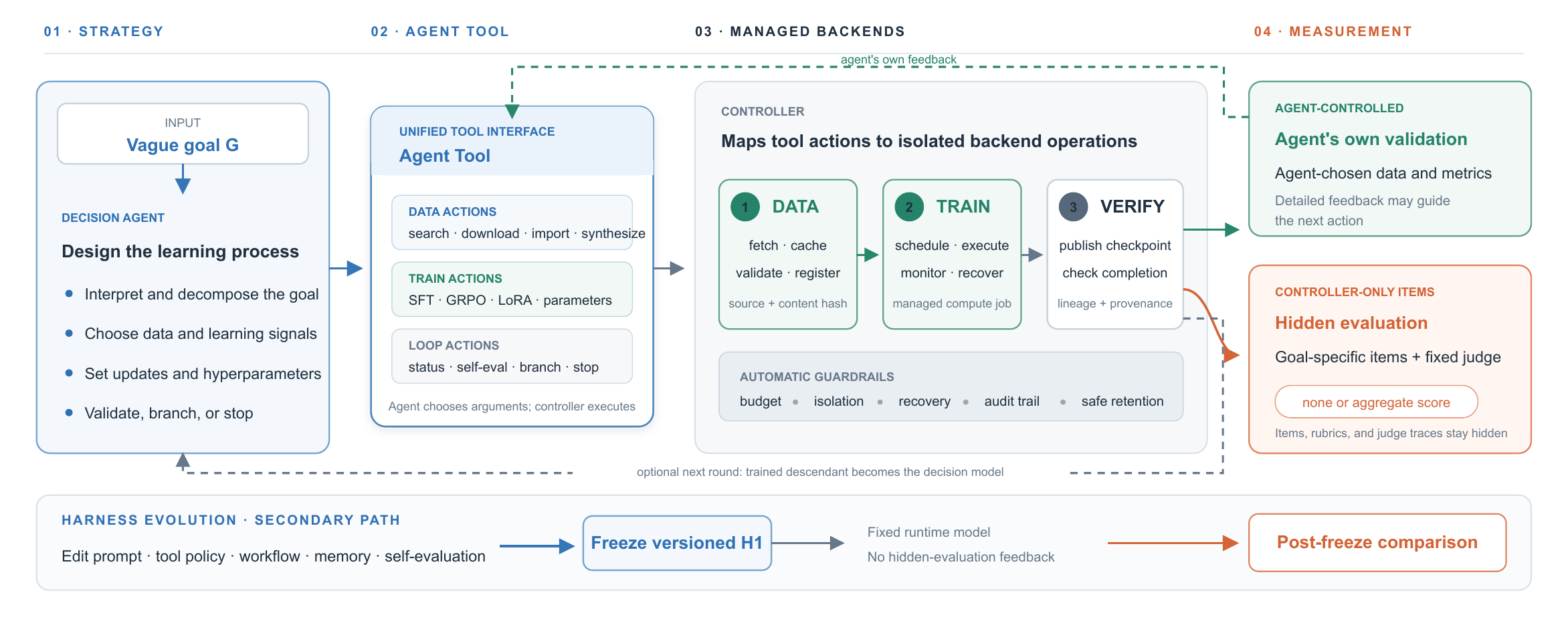}
\caption{The minimal interactive environment and outcome protocols. Weight
candidates follow a data--train--verify path, whereas harness candidates follow
an edit--freeze path. A unified agent tool exposes composable data, training,
status, and validation actions on the agent's own data; the controller maps
them to managed backends and enforces budget and lineage constraints.
Evaluation returns only the aggregate outcome permitted by the protocol.
Across rounds, the controller may reuse the current decision model or
explicitly promote a verified trained descendant; any handoff begins only in
the subsequent round.}
\label{fig:system-design}
\end{figure*}

\paragraph{An agent-facing execution tool.}
The agent does not operate a shell, download a training repository, assemble a
runtime, or implement distributed execution. Instead, \sysname{} exposes a
single agent tool with typed, composable actions
(Figure~\ref{fig:system-design}). Through tool calls, the agent
can search for, download, or import public datasets; synthesize and register new
data; launch supervised fine-tuning (SFT) or Group Relative Policy Optimization
(GRPO) with low-rank adaptation (LoRA) or other permitted configurations; query
job state; construct and run checks on its own validation data; and branch or
stop a lineage. The agent still chooses the data, update method, parameters,
and control flow, while the controller translates those choices into managed
backend operations. This design minimizes incidental engineering work without
removing the strategic decisions that self-evolution from a vague goal is
intended to test.

\paragraph{Minimal actions and trust boundary.}
For weight evolution, the agent chooses data, update method, hyperparameters,
parent checkpoint, and whether to continue, branch, or stop. For harness
evolution, it edits the runtime prompt, tool policy, workflow, memory, or
validation procedure using its own data and freezes a versioned candidate. The
interface exposes only semantic actions---registration, training or editing,
verification, validation, evaluation, branching, and termination---while
hiding backend mechanics. It preserves the information boundary defined in
Section~\ref{sec:ood-benchmark}: the agent makes strategic decisions, the
controller executes and constrains them, and evaluation items remain hidden.

\paragraph{Verified candidates and auditable execution.}
A weight candidate is evaluable only after the controller verifies its parent,
registered data, admitted update, and complete checkpoint publication. A
harness candidate must be versioned and frozen before evaluation. Invalid,
incomplete, foreign, or agent-invented artifacts cannot enter the scored
lineage. Accepted actions record provenance, candidate lineage, resource use,
aggregate submissions, and terminal selection. Full action schemas, recovery,
and backend implementation details are part of the system artifact; the
experiment-facing accounting rules appear in Appendix~\ref{app:results}.

\paragraph{Candidate outcomes and safe retention.}
Let $Y^{\mathrm{in}}$ be the incoming state and $\mathcal K$ the verified
candidates produced in the round. For candidate $k$, we define the raw score
change as
\[
  \Delta^{\mathrm{raw}}(k)=\widehat s_{\Gamma}(Y(k))
       -\widehat s_{\Gamma}(Y^{\mathrm{in}}),
\]
which may be positive or negative. The \emph{best evaluated checkpoint} is the
highest-scoring evaluated checkpoint; the \emph{selected checkpoint} is the
highest-scoring eligible checkpoint under the campaign's predeclared
eligibility and completion rules. For score-gated weight evolution, the
controller retains the selected checkpoint $k^\star$ only when
$k^\star\neq\bot$ and $\Delta^{\mathrm{raw}}(k^\star)>0$; otherwise it rolls
back to $Y^{\mathrm{in}}$. The resulting state $Y^{\mathrm{out}}$ is therefore
$Y(k^\star)$ in the former case and $Y^{\mathrm{in}}$ otherwise. Consequently,
\[
  \Delta^{\mathrm{ret}}
  =\widehat s_{\Gamma}(Y^{\mathrm{out}})
   -\widehat s_{\Gamma}(Y^{\mathrm{in}})\geq0.
\]
The nonnegative retained improvement follows from safety rollback and is therefore a
selection outcome, not evidence that every attempted update improves
capability. Harness candidates are instead compared post-freeze with model
weights fixed; their effect may be negative and does not imply deployment or
recursive handoff. Full eligibility, tie-breaking, termination, and aggregation
rules for the reported experiments appear in
Appendix~\ref{app:outcome-accounting}.

\section{Experiments and Analysis}
\label{sec:experiments}

\rev{We organize the experiments around the three questions introduced in
Section~\ref{sec:intro}. \textbf{RQ1} asks how vague goals change
post-training outcomes and search trajectories relative to an explicit-task setting.
\textbf{RQ2} asks whether an LLM can turn a vague goal into a capability gain
through weight updates. \textbf{RQ3} asks whether the agent harness can
itself evolve when runtime model weights are fixed. Together, the three studies
move from characterizing the effect of goal specification, to evolving the
model, and finally to evolving the system around it.}

\subsection{Setup}
\label{sec:setup}

\rev{RQ1 changes goal specification and uses the original PostTrainBench
evaluator. RQ2 uses two complementary protocols over six vague goals: the
final-only protocol comprises 24 runs without intermediate evaluation scores,
with two terminal runs per model--goal pair reported as their arithmetic mean
(Avg@2); the adaptive-feedback protocol comprises 30 configuration--goal cells
and tests feedback-guided checkpoint search. RQ3 changes the harness while
fixing the runtime model and hidden evaluation set.}

\rev{We report the two RQ2 protocols separately. In RQ3, three executions of
each frozen harness measure runtime variation, not three independent evolution
runs. Detailed budgets, evaluator settings, and outcome definitions appear in
Appendix~\ref{app:results}.}

\subsection{RQ1: How Vague Goals Change Post-Training}
\label{sec:rq1}

\paragraph{Question.} \rev{How do vague goals change post-training outcomes
and search trajectories relative to an explicit-task setting?}

\paragraph{Setting.}
\rev{Standard PostTrainBench names the downstream benchmark, thereby specifying
what the agent should optimize. We replace that identifier with a broad
capability description while preserving the post-training interface and original
task evaluator. The resulting agent must choose, explicitly or implicitly, a
concrete goal, a proxy training task, and a validation method in addition to
the update itself. We use the released PostTrainBench scores as a reference and
matched Claude Opus 4.8 trajectories under explicit-task and vague-goal
contracts only for process comparisons. Other run
metadata can also differ, so the results below describe the observed difference
rather than isolate a prompt-only causal effect. Tasks, models, and pairing
criteria are given in Appendix~\ref{app:rq1-details}.}

\begin{figure*}[t]
\centering
\includegraphics[width=\textwidth]{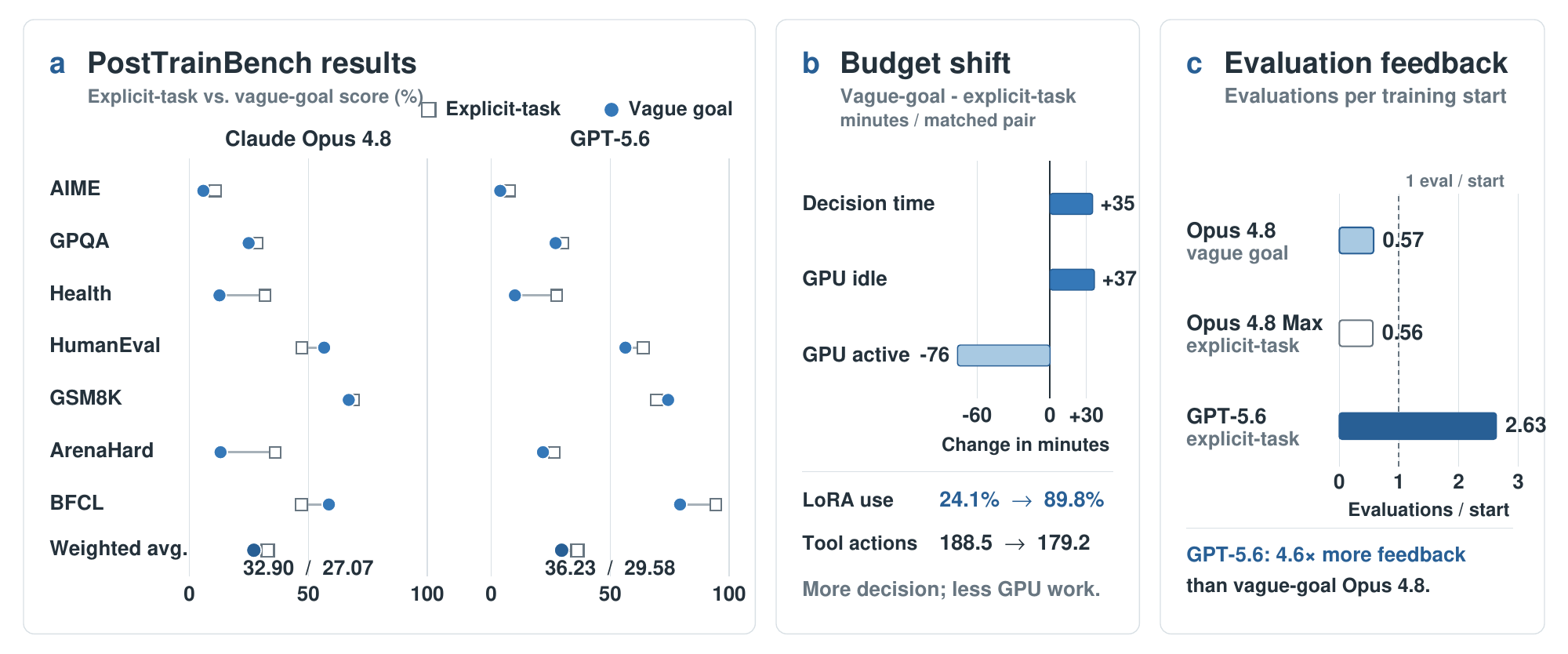}
\caption{RQ1 final-score and trajectory comparison. (a) Official
PostTrainBench references for Claude Opus 4.8 Max and GPT-5.6, alongside
vague-goal scores across seven benchmarks. (b)
Within matched Claude Opus 4.8 pairs, vague-goal trajectories show more
decision and GPU-idle time and less active GPU work. (c) Claude Opus 4.8 under
vague-goal prompting obtains less evaluation feedback per training start
than official GPT-5.6, while official Claude Opus 4.8 Max has
nearly the same feedback density. Panel (c) is a descriptive
cross-system reference, not a matched prompt effect. Exact values
and accounting appear in Appendix~\ref{app:rq1-details}.}
\label{fig:rq1-result}
\end{figure*}

\paragraph{Main result: vague-goal final scores are lower overall, but
task-level changes vary.}
\rev{Figure~\ref{fig:rq1-result}(a) shows that Claude Opus 4.8 under
vague-goal prompting reaches 27.07, compared with the 32.90 official Claude
Opus 4.8 Max reference from PostTrainBench; GPT-5.6 reaches 29.58, compared
with the 36.23 official GPT-5.6 reference. Yet the task-level pattern varies:
Claude Opus 4.8 under vague-goal prompting is higher on HumanEval and BFCL,
and GPT-5.6 under vague-goal prompting is higher on GSM8K. We
therefore interpret the aggregate gaps without assuming uniform task-level
degradation.}

\paragraph{Trajectory pattern 1: vague goals are associated with different
resource use.}
\rev{In the matched Claude Opus 4.8 runs, vague-goal trajectories record 2,109
more seconds of decision-model thinking and 0.61 more GPU-idle hours per
matched run pair, while active training and evaluation time falls by 1.27 hours
(Figure~\ref{fig:rq1-result}(b)). Accesses to agent-visible task materials and
proxy evaluation scripts become $2.98\times$ and $2.39\times$ as dense. In a
separate trace-level sample spanning the six shared settings, the mean number
of tool actions does not increase (188.5 to 179.2), while LoRA is used more
often under vague goals. These descriptive shifts are consistent with more
goal-operationalization work and less active update time, but do not establish
that one displaces the other or identify which choice causes the final score
gap.}

\paragraph{Reference comparison: strong explicit-task systems use different
search profiles.}
\rev{Claude Opus 4.8 under vague-goal prompting performs 3.54 evaluations for 6.17 training starts,
or 0.57 evaluations per start. Official GPT-5.6 performs 39.21 evaluations for
14.91 starts, or 2.63 per start---$4.6\times$ the feedback density
(Figure~\ref{fig:rq1-result}(c)), providing more opportunities to test, reject,
and branch from updates. Official Claude Opus 4.8 Max follows a different profile: its 2.89
evaluations and 5.14 training starts yield 0.56 evaluations per start, nearly
identical to Claude Opus 4.8 under vague-goal prompting, but it records 165,851 thinking characters versus 30,895
for official GPT-5.6. These cross-system statistics describe two strong
explicit-task profiles rather than isolate goal vagueness: dense evaluation
feedback for official GPT-5.6 and much larger recorded reasoning volume for
official Claude Opus 4.8 Max.}

\paragraph{Answer to RQ1.}
\rev{Relative to explicit-task references, vague goals expand the search from
\emph{how to optimize} to \emph{what to optimize and how}. In the matched
Claude Opus 4.8 runs, vague goals are associated with more goal-operationalization
work and less active update time, while their aggregate scores remain below the
explicit-task references.}

\subsection{RQ2: Self-Directed Weight Evolution from Vague Goals}
\label{sec:rq2}

\paragraph{Question.} \rev{Starting from an instruction-tuned checkpoint,
can an LLM given only a vague goal determine what to learn and improve its
capability by directing iterative updates to that checkpoint? This starting
point more closely reflects practical self-evolution, where a generally useful
model must improve itself while preserving behavior acquired through
instruction tuning.}

\paragraph{Settings.}
\rev{Unlike RQ1, which adapts pretrained models, both RQ2 settings begin from
instruction-tuned checkpoints and allow iterative training. We call each
initial instruction-tuned checkpoint the \emph{base model} and its initial
evaluation score the \emph{base score}. This is a practically important
setting: an update should add goal-relevant behavior without disrupting useful
behavior already encoded by instruction tuning. The final-only protocol tests
open-loop execution, whereas the adaptive-feedback protocol tests search with
sparse aggregate scores. Although the system can promote a trained descendant to the
decision model in a subsequent round, all reported runs keep the initial
decision model fixed; the evidence therefore concerns adaptive weight search
rather than recursive decision-model replacement. In both protocols,
evaluation focuses on one goal and does not measure preservation of every
unrelated capability.}

\paragraph{Final-only protocol.}
\rev{Qwen3.5-4B Self and Qwen3.5-9B Self each optimize all
six goals in two separately executed terminal runs, yielding 24 runs in total.
For each model--goal pair, we report Avg@2---the arithmetic mean of the two
final-checkpoint scores---rather than best-of-two. A run may launch multiple
training jobs within 40 GPU-hours, but it may submit only one final checkpoint
for evaluation on the hidden evaluation set. The two-run repetition makes the
final estimate less dependent on any single run. This protocol does not isolate
decision-model scale because the decision model, the base model, and the base
score change together between 4B and 9B.}

\begin{figure}[t]
\centering
\includegraphics[width=\linewidth]{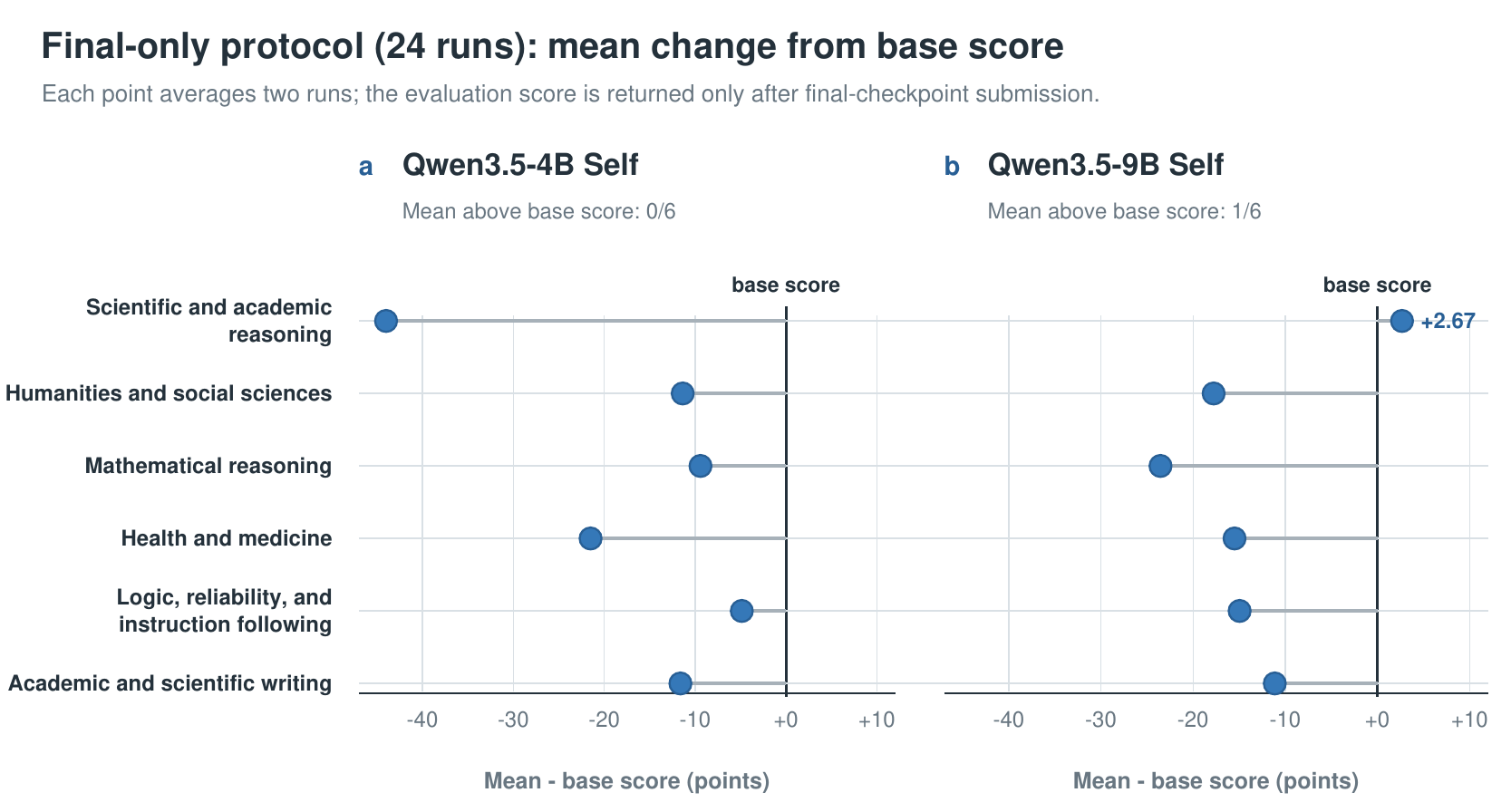}
\caption{Final-only mean change from the base score before rollback. Each
point is the arithmetic mean of two separately executed terminal runs for
one model--goal pair. Scientific and academic reasoning with Qwen3.5-9B is the
only two-run mean above its base score. Constituent run scores appear in
Appendix~\ref{app:oneshot24}.}
\label{fig:oneshot24}
\end{figure}

\paragraph{Final-only result: only one two-run mean exceeds the base score.}
\rev{All 24 final-only runs complete their single final-checkpoint evaluation
(Figure~\ref{fig:oneshot24}). Taking each model--goal pair as the reporting
unit, Qwen3.5-4B has 0/6 mean outcomes above its base score and Qwen3.5-9B has 1/6.
The sole positive pair is scientific and academic reasoning: both runs rise
from 45.33 to 48.00, so the two-run mean is 48.00, or 2.67 points above the base score. At the
pair level, this is the only positive mean among all 12 model--goal pairs.
At the individual-run level, 3/24 final checkpoints exceed their base scores,
while rollback retains the base model in the other 21 runs.
Individual terminal-run and per-goal results appear in
Appendix~\ref{app:oneshot24}.}

\paragraph{Adaptive-feedback protocol.}
\rev{The adaptive-feedback protocol asks whether sparse aggregate evaluation scores can guide an
improving multi-checkpoint trajectory. The decision model chooses public data,
an update method, a training plan, and when to evaluate or branch in response
to the scores observed so far. In the Self settings, the initial Qwen3.5-4B or
Qwen3.5-9B checkpoint remains the fixed decision model while descendant
checkpoints are trained. GPT-5.6 Luna, Terra, and Sol serve as external
decision models for the same Qwen3.5-4B base model.
Eligible checkpoints may receive repeated aggregate scores on the hidden
evaluation items for a goal, but the agent never sees the items or per-item
feedback.}

\begin{figure}[t]
\centering
\includegraphics[width=\linewidth]{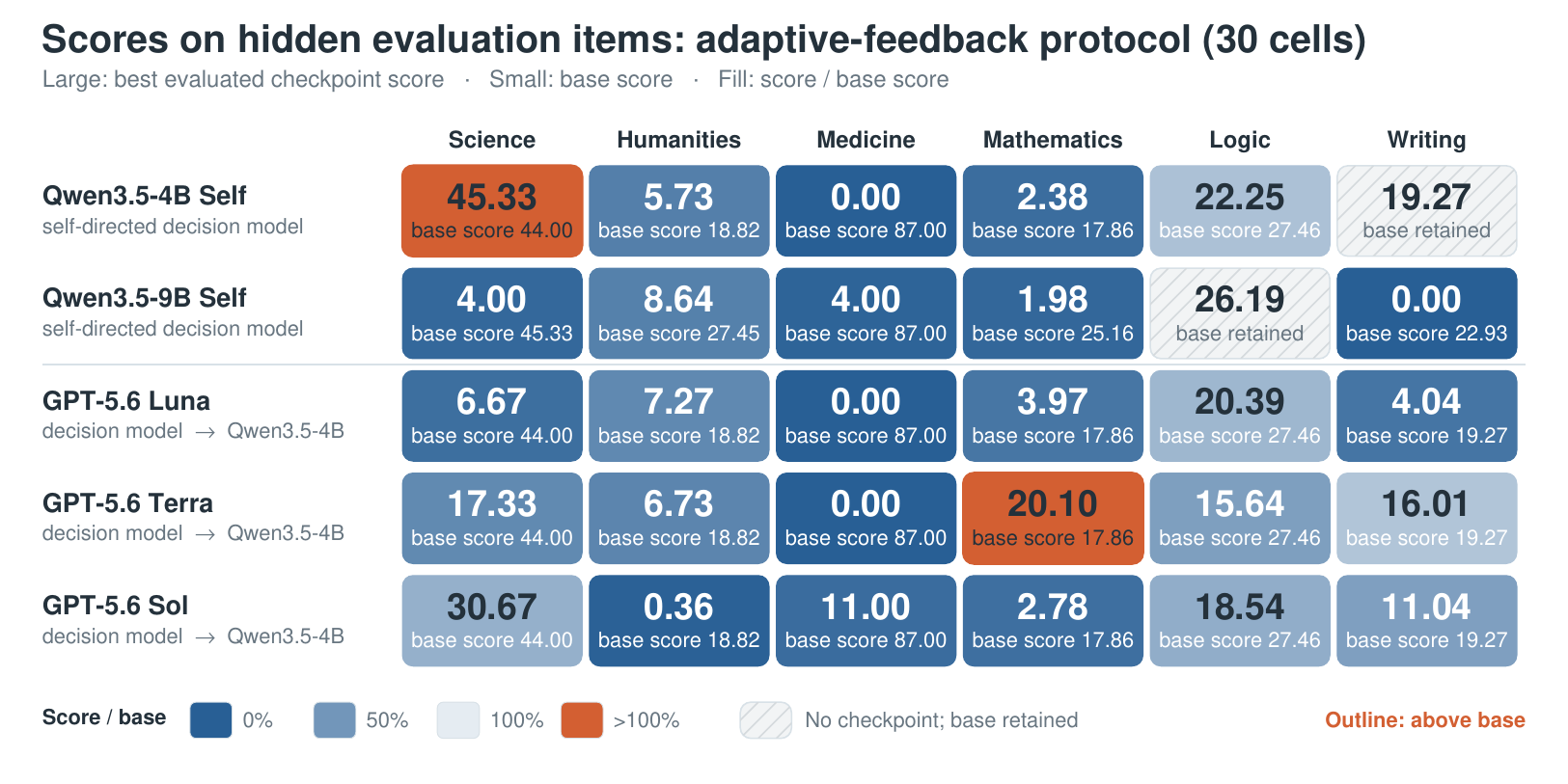}
\caption{Scores on the hidden evaluation set for all 30 adaptive-feedback
configuration--goal cells: two Self configurations and three external decision-model settings. Large
values are the scores of the best evaluated checkpoints; small values give the base
score. The two cells without an evaluated checkpoint show the fallback base
score for accounting but are hatched and labeled as having no checkpoint. For
evaluated cells, color encodes the best evaluated checkpoint score divided by
the base score. Printed values preserve the original score scale.}
\label{fig:rq2-outcome-map}
\end{figure}

\begin{figure}[t]
\centering
\includegraphics[width=\linewidth]{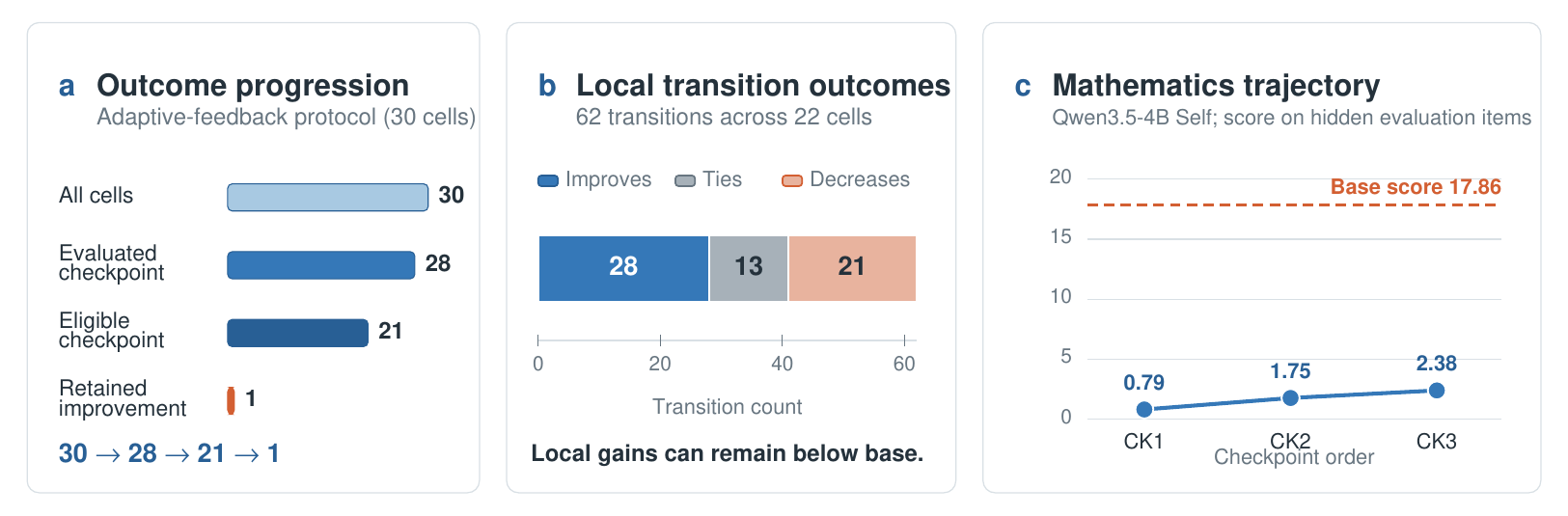}
\caption{Adaptive-feedback outcome and trajectory decomposition. Of 30 cells,
28 produce an evaluated checkpoint, 21 produce an eligible checkpoint, and one
yields a retained improvement (a). Of 62 consecutive within-round checkpoint transitions, 28
increase (b),
but even a monotonic self-optimization trajectory can remain far below the
base score (c).}
\label{fig:rq2-trajectory}
\end{figure}

\paragraph{Adaptive-search result: producing a trained checkpoint is common;
producing an above-base goal score is rare.}
\rev{Across the 30 configuration--goal cells, with one run per
configuration--goal cell, 28 produce an evaluated checkpoint and 21
satisfy the predeclared provenance, exploration, completion, and
terminal-record requirements for an eligible checkpoint
(Figures~\ref{fig:rq2-outcome-map} and~\ref{fig:rq2-trajectory}(a)). Only two
cells have a best evaluated checkpoint whose score exceeds the base score. Qwen3.5-4B Self
raises science from 44.00 to 45.33, but no Self cell retains a model above its
base score; Terra raises
Qwen3.5-4B mathematics from 17.86 to 20.10 and is the only cell whose
checkpoint scores above the base score, passes eligibility, and remains after
rollback. Thus the agents often close the executable training loop---data,
update, checkpoint, and evaluation---but most resulting checkpoints do not
preserve the base score. The 45.33 science score is a
best evaluated checkpoint from one adaptive-feedback run and is not reproduced by
the two final-only Qwen3.5-4B final checkpoints. Because feedback and
checkpoint-selection contracts differ, we treat the protocols as complementary
settings rather than repeated trials. The retained Terra score was
selected using repeated aggregate feedback on the same fixed evaluation slice;
without a separate confirmation slice, it is evidence of feedback-guided
selection rather than an independently replicated capability gain.}

\paragraph{Trajectory pattern 1: apparent progress can be recovery from training-induced
regression.}
\rev{The 22 cells with multiple evaluated checkpoints contain 62 consecutive
transitions: 28 rise, 13 tie, and 21 fall. Later search beats the first evaluated
checkpoint in 14/22 cells, but this does not imply a score above the base score.
For example, Qwen3.5-4B Self improves on mathematics from 0.79 to 1.75 to 2.38
while remaining far below its 17.86 base score
(Figure~\ref{fig:rq2-trajectory}(c)). A positive within-lineage slope can
therefore describe recovery from training-induced regression rather than an
above-base improvement. The transition counts are pooled descriptive statistics because
cells contribute different numbers of checkpoints.}

\begin{figure}[!t]
\centering
\includegraphics[width=0.88\linewidth]{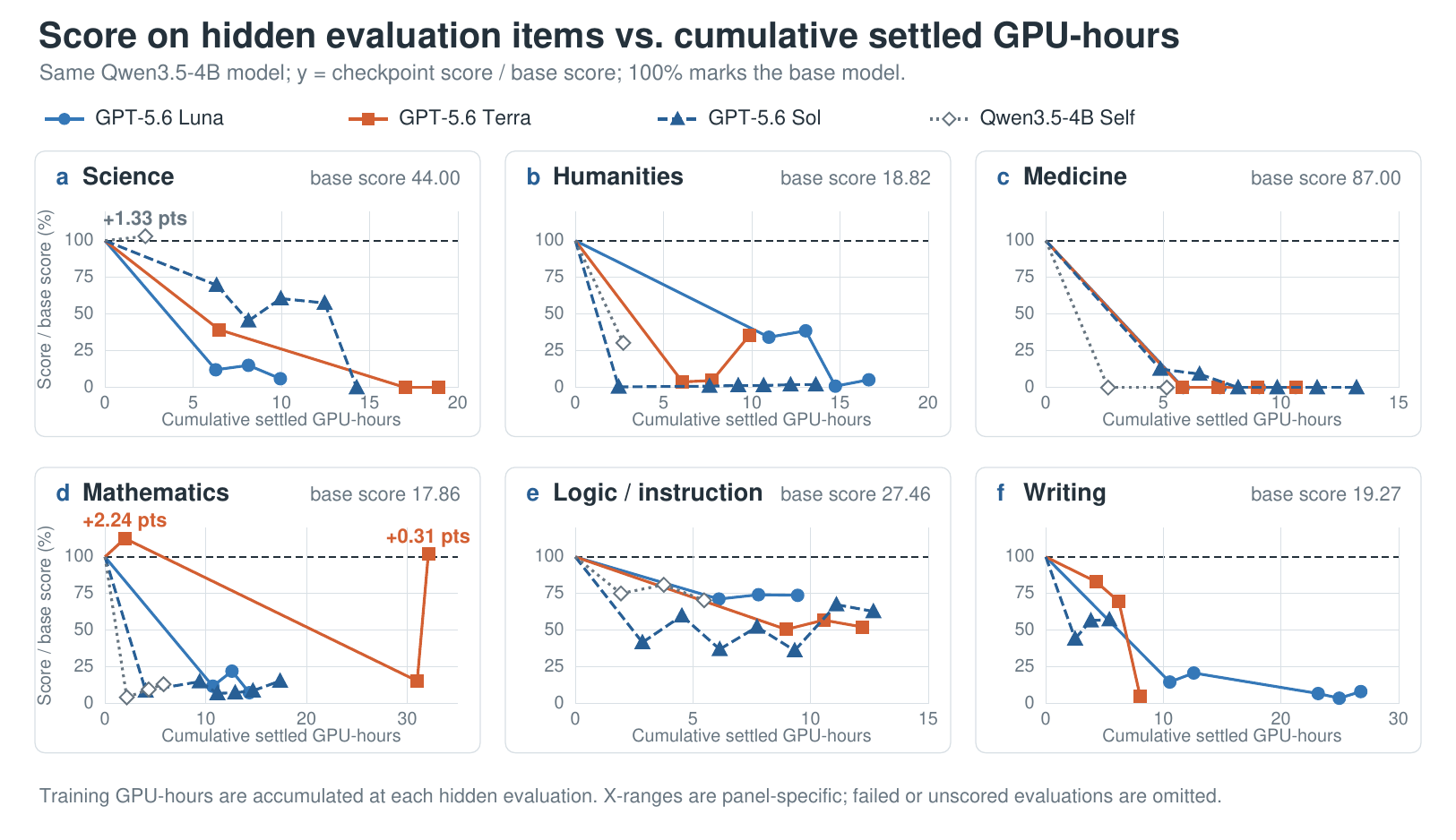}
\caption{Adaptive-feedback score--compute trajectories for four configurations
using the same Qwen3.5-4B base model. Scores are normalized to each goal's base
score, and curves connect evaluated checkpoints by settled training GPU-hours.}
\label{fig:rq2-gpu-score}
\end{figure}

\paragraph{Trajectory pattern 2: more search does not mean better search.}
\rev{Using the same Qwen3.5-4B base model, the external decision models produce
different goal-level outcomes. Among their best evaluated checkpoints, Luna is highest on humanities
(7.27) and logic (20.39), Terra on mathematics (20.10) and writing (16.01), and
Sol on science (30.67) and medicine (11.00). In five of six goals, even the
highest of these scores remains below the base score. Sol searches most broadly,
producing 33 evaluated checkpoints with 76.56 settled training GPU-hours, yet
none exceeds its corresponding base score. Terra finds the sole retained
improvement on the adaptive selection slice; its later mathematics descendants
fall from 20.10 to 2.78 before
recovering to 18.17. Greater search volume therefore does not ensure an
above-base score, and continued optimization can induce regression rather than
produce a better descendant, making checkpoint selection and rollback necessary
(Figure~\ref{fig:rq2-gpu-score}).}

\rev{In the final-only protocol, three runs continue for 16--20 training jobs
and consume 26.570--32.286
GPU-hours, yet their submitted final checkpoints score only 0.159 on 4B mathematics,
2.540 on 9B mathematics, and 0 on 9B writing. Because intermediate checkpoints
are not evaluated under this protocol, these trajectories do not establish monotonic
degradation or identify the best earlier checkpoint; they show that blind
continuation along a latest-checkpoint lineage can end in a poor final
checkpoint.}

\paragraph{Observed failure mode: numeric-label SFT is associated with answer-format
collapse.}
\rev{The Self trajectories show concentration in both domain and answer format.
In the adaptive-feedback protocol, Qwen3.5-4B Self uses GSM8K or Hendrycks mathematics data in 30/32
dataset-import events, including runs targeting science, logic, and writing. In
the final-only protocol, five submitted checkpoints are trained on numeric-label MMLU SFT:
all 21,000 training targets are single-digit labels, and all 279 corresponding
evaluation outputs are single digits. Their scores are 0, 0, 0, 6.141, and
0. All five checkpoints show the same pattern. This is not a randomized causal
comparison, but it documents a concrete answer-format mismatch associated with
training-induced regression. Because all 24 final evaluations complete
without item errors, missing outputs or evaluator failures do not explain the
aggregate degradation. Further accounting appears in
Appendix~\ref{app:oneshot24}; content-dependent interpretations use the
controlled-access source traces.}

\paragraph{Answer to RQ2.}
\rev{We observe isolated above-base goal scores rather than reliable improvement.
Only one of 12 final-only Avg@2 outcomes exceeds its base score, and adaptive
feedback often produces locally rising trajectories that remain below base.
Agents therefore complete the update loop more consistently than they preserve
base performance; these goal-specific evaluations do not measure unrelated
capabilities.}

\FloatBarrier
\subsection{RQ3: Harness Evolution}
\label{sec:rq3}

\paragraph{Question.} \rev{With model weights held fixed, can one-step harness
evolution improve the agent's research performance for the academic and
scientific writing goal?}

\paragraph{Design.} \rev{We fix Qwen3.5-4B as the runtime model and use the
original Qwen-Agent as the reference harness. Under the same editable contract,
Qwen3.5-4B Creator and GPT-5.6 Luna, Terra, and Sol each attempt to produce one
successor harness $H_1$ from the same $H_0$. The Qwen3.5-4B Creator produces no
valid $H_1$ and is assigned zero by the predeclared failure rule. Each of the
three valid successors is frozen and executed three times on the same 20 hidden
evaluation items for the academic and scientific writing goal; those scores are not returned to the
creator. There is one creation trajectory per condition, while the three
executions measure runtime variation of a frozen harness. No run evolves $H_1$
into $H_2$, so the experiment evaluates one-step harness generation rather than
recursive harness self-improvement.}

\paragraph{Result: the closest successor harness is numerically below the fixed
reference.} \rev{All three valid successors are numerically below the original Qwen-Agent, which obtains
28.64 task macro and 27.65 example micro (Table~\ref{tab:rq3}). The Sol-created
successor is closest at 27.22/25.97, 1.42 task-macro points below the reference,
followed by the Terra-created successor at 20.76/20.14 and the Luna-created successor at
19.32/18.33.}

\begin{table}[t]
\centering
\small
\caption{RQ3: research performance on the hidden evaluation set for the fixed Qwen-Agent reference
and successor harnesses. All executable harnesses use Qwen3.5-4B as the runtime
model.}
\label{tab:rq3}
\begin{tabular}{@{}lcc@{}}
\toprule
\textbf{Configuration} & \textbf{Task macro} & \textbf{Example micro} \\
\midrule
Original Qwen-Agent reference      & 28.64 & 27.65 \\
Qwen3.5-4B Creator (no valid harness) & N/A & N/A \\
GPT-5.6 Luna successor harness       & 19.32 & 18.33 \\
GPT-5.6 Terra successor harness      & 20.76 & 20.14 \\
GPT-5.6 Sol successor harness        & 27.22 & 25.97 \\
\bottomrule
\end{tabular}
\vspace{2pt}

\parbox{0.96\linewidth}{\footnotesize\emph{Note:} The reference and each valid
successor are frozen and executed three times on the same 20 evaluation items; their
reported values are execution means. Task macro weights evaluation items equally,
whereas example micro weights evaluator-scored examples equally. The Qwen3.5-4B Creator produces no
valid harness, so it has no execution mean; its predeclared protocol-level
failure score is zero. Per-execution dispersion is not reported, so differences
between executable harnesses are descriptive.}
\end{table}

\paragraph{Trajectory pattern 1: narrow agent validation can specialize the harness to
the wrong proxy.}
\rev{Luna raises one creator-designed eight-item checklist on the agent's own validation data
from 7/8 to 8/8 by adding a fixed five-part research-answer template. After two
similarly structured prompts also score 8/8, it stops after approximately 40
minutes. The template makes answers more explicit about bias and limitations,
but inspection of hidden evaluation items shows that it can recast requests for mechanisms,
formulas, parameters, or engineering details as study-design problems. The
validation gain and the 19.32 task-macro mean on the hidden evaluation set are therefore consistent
with specialization to a narrow proxy rather than broad improvement.}

\paragraph{Trajectory pattern 2: answer review does not guarantee output integrity.}
\rev{Terra correctly rejects a two-pass reviewer after it invents unsupported
significance results and effect sizes, but its final harness retains a different
failure path: after a calculator call, an empty final model response causes the
pre-tool partial draft to be submitted. This occurs on two items in one
inspected execution and accounts arithmetically for approximately 70\% of that
execution's aggregate decline. Sol searches for 3 hours 11 minutes across four
task types in its own validation data and several prompt and tool variants, rejecting alternatives with
hallucination, latency, or tool-control failures even when their validation scores
are higher. Its final edit only checks that the last model response is complete
and nonempty; the check does not trigger on any of the 20 inspected evaluation
items. It therefore improves failure handling, but the trace does not show that
it causes a task-accuracy gain. Detailed cases appear in
Appendix~\ref{app:rq3-traces}.}

\paragraph{Answer to RQ3.} \rev{One-step harness evolution can produce
executable and behaviorally distinct successor harnesses, but all three valid
successors remain numerically below Qwen-Agent. Their trajectories expose
bottlenecks in validation diversity and output integrity: in this one-step,
single-goal setting, agents complete the edit--test loop without improving mean
hidden-set performance over the reference harness.}

\section{Related Work}
\label{sec:related}

\paragraph{Self-training and self-evolution methods.}
Self-evolution methods let models construct parts of their own learning signal.
STaR~\citep{star2022} fine-tunes on self-generated rationales that yield correct
answers; ReST-EM~\citep{restem2023} iterates sampling, reward filtering, and
retraining; Self-Rewarding LM~\citep{selfreward2024} uses the model as judge;
and SPIN~\citep{spin2024} learns through self-play. R-Zero~\citep{rzero2025}
and Absolute Zero Reasoner~\citep{azr2025} further reduce dependence on external
data by posing, solving, and refining problems, while SEAL~\citep{seal2025}
reinforces model-generated self-edits according to downstream gains. In all
cases, the model helps generate data, rewards, tasks, or edits, but progress is
still operationalized by a concrete signal such as correctness, reward,
self-play outcome, or task gain. \sysname{} is complementary: it introduces no
new training algorithm, but tests whether a model given only a vague goal can
construct an effective learning process around it.

\paragraph{Automated post-training and ML-agent benchmarks.}
PostTrainBench~\citep{posttrainbench} gives an agent a base model, evaluation
script, and budget for data filtering, training, and hyperparameter search;
LaMDAgent~\citep{lamdagent2025} similarly automates iterative post-training
pipelines. RSIBench-Data~\citep{meng2026rsibenchdatabenchmarkingdatacentricresearch}
fixes the target model, benchmark, and training stack while agents revise data
strategies, and finds that most continued searches finish below their own best
checkpoint---a retention failure related to our distinction between candidate
progress and retained gain. AI4AI-Bench~\citep{chi2026ai4aibenchbenchmarkingllmagents}
instead lets agents rewrite a training algorithm under hidden evaluation, yet
most submissions do not change how the model learns. PAST-Bench~\citep{xue2026pastbenchbenchmarkingfoundationsrecursive}
tests whether retained experience improves later personal-agent tasks through
the intended save--retrieve--update pathway.

Broader benchmarks such as MLE-bench~\citep{mlebench2024},
MLAgentBench~\citep{mlagentbench2024}, and RE-Bench~\citep{rebench2024} measure
ML experimentation and AI R\&D against quantifiable objectives. AIDE~\citep{aide2025},
DS-Agent~\citep{dsagent2024}, and SELA~\citep{sela2024} automate data-science
and modeling pipelines; ADAS~\citep{adas2024} and Darwin G\"odel
Machine~\citep{dgm2025} improve agent systems or code. Hyperagents~\citep{zhang2026hyperagents}
makes the meta-level modification procedure editable, Frontis-MA1~\citep{yang2026frontisma1trainingai4aimodel}
trains a dedicated meta-evolution agent, and the Red Queen G\"odel
Machine~\citep{iacob2026redqueengodelmachine} co-evolves agents and evaluators
so the objective changes across epochs.

These systems make automated engineering measurable, but generally start from
an already operationalized task, score, or evaluator. Complementary shadow
evaluations find that frontier agents can complete engineering without advancing
an unpublished research question~\citep{kirgis2026aiagentsconductopenended},
and a survey of 1,250 self-improvement papers identifies research
direction-setting as the bottleneck that keeps humans in the loop~\citep{chen2026recursiveselfimprovementaibounded}.
This matters when an existing metric saturates or no longer represents the next
capability gap. \sysname{} withholds the downstream task and measures whether an
agent can turn a broad direction into learning measured on hidden evaluation
items: RQ1 isolates this goal-side difference within PostTrainBench, and RQ2 studies weight
evolution directed by external models or the base model itself.

\paragraph{Memory, workflow, and skill evolution.}
Self-improvement can also occur through memory, tools, and workflows.
Evo-Memory~\citep{evomemory2025} studies writing, organizing, and reusing memory;
EvoFSM~\citep{evofsm2026} uses interpretable finite-state workflow edits; and
SkillLearnBench~\citep{zhong2026skilllearnbenchbenchmarkingcontinuallearning}
finds that external feedback supports continual skill gains whereas
self-feedback alone can cause recursive drift. HELIX~\citep{fan2026helixmodelharnesscoevolutionrecursive}
rebuilds a fixed model's harness, updates the model from verified trajectories,
and rebuilds the harness again; Co-Harness~\citep{chen2026coharnesscoevolvingharnessesmodel}
alternates validated harness edits with fine-tuning on trajectories produced by
the improved harness. RQ3 connects this line to vague-goal learning by asking
whether the harness mediating goal interpretation, tool use, memory, and
self-evaluation can evolve while runtime weights remain fixed.

\section{Conclusion}
\label{sec:conclusion}
\enlargethispage{2\baselineskip}

Modern LLM agents can search for effective ways to improve a specified
objective. The next step is to determine what should be learned when no
agent-visible task specification or decomposable reward has already
operationalized progress. We formalize this problem as vague-goal-driven
self-evolution and introduce \sysname{}, which
pairs vague goals with hidden evaluation and
auditable trajectories. Across three studies, the RQ1 process slice associates
vague-goal prompting with more goal-definition work; self-directed
runs produce evaluated checkpoints far more often than retained improvements; and the
highest successor-harness mean remains below a fixed reference under
unchanged runtime weights. These
results show why progress must be judged against the base model rather
than only against the previous checkpoint: closing the training loop is not yet
the same as closing the capability loop.

Our conclusions are bounded by the six goals and the coverage and scoring
quality of their expert-authored evaluation items, one canonical
adaptive-feedback run per configuration--goal cell, and content-level trace
evidence available only under controlled access. Broader goal coverage,
repeated runs, refreshed evaluation-set versions, and
recursive evolution that controls drift and unrelated capability loss remain
important directions for future work.

\clearpage
\section{Contributions}

\textbf{Core Contributors} \\
Yuhao Wu, Jingyuan Zhang, Jiajun Shi

\vspace{0.5em}

\textbf{Contributors} \\
Yuxuan Zhang, Xinping Lei, Junting Zhou, Zexuan Wang, Yuchen Wu,
Huan Zhou, Duo Wang, Yinzhu Piao, Yongchang Peng, Yunfeng Shi, Jin Chen,
Zuo Wang, Jinkai Liu, Jiaheng Liu

\vspace{0.5em}

\textbf{Corresponding Authors} \\
Yuhao Wu (\email{wuyuhao.621@bytedance.com})\\
Wenxuan Zhang (\email{wxzhang@sutd.edu.sg})\\
Shen Yan (\email{sheny@bytedance.com})\\
Wenhao Huang (\email{huang.wenhao@bytedance.com})\\
Ge Zhang (\email{gezhang@umich.edu})

\bibliographystyle{plainnat}
\bibliography{main}

\clearpage
\begingroup
\setlength{\parskip}{0.25\baselineskip plus 1pt minus 0.5pt}
\beginappendix

\section{Benchmark Construction and Experimental Contracts}
\label{app:benchmark-details}

\subsection{Goal and Evaluation-Set Statistics}
\label{app:goals}

\rev{The benchmark defines six vague goals, each paired with one mutually
exclusive group of items in the hidden evaluation set
(Table~\ref{tab:goal-map}). The adaptive-feedback protocol evaluates all six
goals. Logic, reliability, and instruction following form one composite
goal in the current release; we do not treat its three components as separately
identified goals.}

\begin{table}[!ht]
\centering
\small
\caption{The six vague goals and their corresponding evaluation items. All 520
top-level items are newly authored by domain experts.}
\label{tab:goal-map}
\resizebox{\linewidth}{!}{%
\begin{tabular}{@{}llr@{}}
\toprule
\textbf{Vague goal} & \textbf{Evaluation items} & \textbf{\#Items} \\
\midrule
Scientific and academic reasoning & Science and academic reasoning & 75 \\
Humanities and social-science knowledge & Humanities and social sciences & 110 \\
Health and medical reasoning & Health and medicine & 100 \\
Mathematical reasoning & Mathematical reasoning & 126 \\
Logic, reliability, and instruction following & Logic, reliability, and instruction following & 89 \\
Academic and scientific writing & Academic and scientific writing & 20 \\
\midrule
\textbf{Total unique items} & \textbf{Six non-overlapping groups} & \textbf{520} \\
\bottomrule
\end{tabular}}
\end{table}

\rev{The items for the academic and scientific writing goal comprise 20
top-level task bundles, counted as 20 evaluation items. Its harness evaluator may produce multiple
scored examples within a task, which is why RQ3 reports both task-macro and
example-micro aggregation.}

\subsection{Contamination-Control Details}
\label{app:contamination}

\rev{Contamination control begins at authorship. Domain experts write each item from
scratch, while named public benchmarks serve only as task-archetype and
difficulty references. Before freezing a version of the hidden evaluation set, the pipeline removes
exact and semantic duplicates within and across reporting groups and audits
overlap against the reference benchmarks. Dataset registration then applies a
second exact and semantic overlap gate against the hidden evaluation set,
and detected matches are rejected before data can enter SFT, GRPO, or
continual-pretraining during a campaign.}

\rev{Seed-2.0, GPT-5.2, and Gemini-3 independently answer each candidate under a
shared blind-screen protocol. L0--L3 denote the number of models that fail;
timeouts and service errors are retried rather than counted as failures. This
screen calibrates difficulty and is separate from the authorship and overlap
audits. After review, scorer binding, and local end-to-end validation, the hidden evaluation set
is frozen under a manifest and SHA-256. Corrections create a new version; a
running campaign continues to use the version fixed at creation.}

\subsection{Prompt Contracts}
\label{app:prompts}

\rev{Prompts are rendered from a versioned setting profile rather than written ad
hoc for each run. A vague-goal prompt contains a natural-language capability
goal, the model being trained, the action contract, budgets, and the rule that
the hidden evaluation set returns aggregate scores only. It does not reveal evaluation item
content, answers, rubrics, goal-routing identifiers, or per-item feedback. For
example, the mathematical goal asks the agent to improve difficult multi-step
reasoning, precise calculation, and clear justification without naming the
evaluation items.}

\rev{An explicit-task contract additionally discloses the approved task format,
difficulty anchor, metric, and success criterion, while preserving the same
content boundary. Every campaign binds its setting identifier, prompt-profile
version, rendered prompt SHA-256, and experiment identity. These bindings are
checked on creation and recovery so that a vague-goal run cannot silently resume
under an explicit-task contract or a different prompt version.}

\section{Complete RQ1 Score and Trajectory Accounting}
\label{app:rq1-details}

\paragraph{Score comparison.}
Table~\ref{tab:rq1-endpoints} gives the exact values visualized in
Figure~\ref{fig:rq1-result}(a). Official scores are system-level PostTrainBench
references; the vague-goal columns are our results. They
share the benchmark definition and aggregation weights, but are not treated as
prompt-only causal pairs.

\begin{table}[!ht]
\centering
\small
\setlength{\tabcolsep}{9pt}
\caption{PostTrainBench score comparison for RQ1. Values are percentages;
the weighted average uses the official benchmark weights.}
\label{tab:rq1-endpoints}
\begin{tabular}{@{}lrrrr@{}}
\toprule
\textbf{Benchmark} & \shortstack{\textbf{Official}\\\textbf{Claude Opus 4.8 Max}} &
\shortstack{\textbf{Vague goal}\\\textbf{Claude Opus 4.8}} &
\shortstack{\textbf{Official}\\\textbf{GPT-5.6}} &
\shortstack{\textbf{Vague goal}\\\textbf{GPT-5.6}} \\
\midrule
AIME 2025   & 10.83 & 5.83  & 7.50  & 3.75 \\
GPQA Main   & 28.35 & 24.89 & 30.13 & 27.04 \\
HealthBench & 31.68 & 12.62 & 27.39 & 9.93 \\
HumanEval   & 47.33 & 56.63 & 63.95 & 56.40 \\
GSM8K       & 68.87 & 66.98 & 69.48 & 74.35 \\
ArenaHard   & 36.09 & 13.07 & 26.40 & 21.72 \\
BFCL        & 47.13 & 58.63 & 94.38 & 79.38 \\
\midrule
\textbf{Weighted average} & \textbf{32.90} & \textbf{27.07} &
\textbf{36.23} & \textbf{29.58} \\
\bottomrule
\end{tabular}
\end{table}

\paragraph{Matched process accounting.}
The process comparison pairs explicit-task and vague-goal runs with the same
decision model, base model, benchmark leaf, and trial index. We exclude
anti-cheat hits and run pairs that return only the fallback base. The matched Claude
Opus 4.8 comparison contains 48 matched run pairs. Each trial occupies one H20 sandbox
GPU for at most 10 hours. We partition occupancy into decision-model waiting
time, during which the sandbox GPU is idle, and active time used by the
remaining workflow.

\rev{For Claude Opus 4.8, vague-goal prompting is associated with 2,109 more seconds of
decision-model thinking and 0.61 more hours of idle time per matched run pair, alongside
1.27 fewer active hours. The corresponding absolute means move from 0.6 to 1.2 idle hours and
from 8.0 to 6.7 active hours. Figure~\ref{fig:rq1-result}(b) converts these
deltas to minutes for a common visual scale; displayed values use the
unrounded measurements.}

\paragraph{Behavioral normalization.}
Narrative traces differ substantially in length across conditions, so action
mentions are normalized per 10,000 characters and interpreted only when hit
rate and density move consistently. Under vague-goal prompting, the
density of reading task definitions rises by $2.98\times$, inspecting
evaluation scripts by $2.39\times$, and low-rank adaptation (LoRA)/parameter-efficient
fine-tuning (PEFT) use by $3.09\times$.
LoRA appears in 89.8\% of vague-goal runs versus 24.1\% of matched
explicit-task runs; 33 matched run pairs switch from not using LoRA to using it, with no switch
in the opposite direction.

We separately validate the Claude Opus 4.8 pattern with trace-level tool calls on the six
matched settings shared by both trace exports (21 trajectories). Mean action count changes
from 188.5 to 179.2 per trajectory. We therefore interpret the longer vague-goal
narrative as additional deliberation, not as evidence that the agent executed
more work. Training starts are counted only from concrete launch commands
(e.g., \texttt{accelerate launch} or \texttt{torchrun}); broad keyword matches
over \texttt{train} and \texttt{sft} are not used as action evidence. These
trace-level calls are broader than the accepted controller actions $a_{r,j}$
defined in Section~\ref{sec:formulation}.

\paragraph{Official-trajectory comparison.}
We apply the same trajectory parser to 112 released official traces, 56 each
for the official Claude Opus 4.8 Max and GPT-5.6 Sol systems. Vague-goal Claude Opus 4.8 averages 3.54 evaluations and
6.17 training starts per trajectory; official Claude Opus 4.8 Max averages 2.89 and 5.14,
while official GPT-5.6 averages 39.21 and 14.91. These correspond to 0.57, 0.56,
and 2.63 evaluations per training start. Official Claude Opus 4.8 Max additionally records
165,851 thinking characters per trajectory, whereas official GPT-5.6 records
30,895, supporting the deep-deliberation and high-frequency-experimentation
interpretations in the main text.

\section{Complete RQ2 Results and Experimental Protocol}
\label{app:results}

\subsection{Outcome Accounting and Rollback}
\label{app:outcome-accounting}

Only checkpoints whose parent, registered data, admitted update, and published
artifacts are verified enter the evaluated lineage. The campaign fixes its
hidden evaluation set, scorer or judge, decoding and aggregation rules, query
allowance, eligibility requirements, terminal selector, and tie-breaking order
before execution.

An \emph{evaluated checkpoint} is any verified candidate with a durable score.
An \emph{eligible checkpoint} additionally satisfies the predeclared
provenance, exploration, completion, and terminal-record requirements. The
\emph{selected checkpoint} is the highest-scoring eligible checkpoint under the
fixed selector and tie-breaking order. The \emph{retained model} uses that
checkpoint only when it exceeds the base score; otherwise the controller rolls
back to the base. We report the best evaluated, selected, and retained outcomes
separately because a high intermediate score may come from an incomplete run,
while a nonnegative retained score may reflect rollback rather than learning.
RQ3 instead freezes each harness before research evaluation, compares it with a
fixed-weight reference, and withholds the score from the creator.

\subsection{Feedback and Submission Protocol}

Vague-goal and explicit-task configurations use separate versioned settings and cannot
submit checkpoints across experiments. Training uses registered data and a
controller-resolved parent, and only a completely published checkpoint from the
current lineage can be validated or submitted.

The agent's own validation data are agent-visible and do not consume queries to
the hidden evaluation set. The adaptive-feedback protocol permits repeated
aggregate-score queries on the same goal-specific evaluation items within its
recorded allowance. The final-only protocol returns no intermediate
score and accepts one final-checkpoint evaluation per run. Invalid or duplicate
requests do not consume a query; an accepted reservation remains counted if a
later deployment or evaluator stage fails. No protocol returns hidden evaluation items,
answers, candidate outputs, rubrics, or judge reasoning.

\subsection{Final-Only Protocol: Vague-Goal Weight Evolution}
\label{app:oneshot24}

\rev{The final-only protocol crosses Qwen3.5-4B Self and Qwen3.5-9B Self with
six goals, each with separately executed Run A and Run B, for 24 runs in total
(Table~\ref{tab:oneshot24}). Each run may launch
multiple training jobs within 40 GPU-hours but may submit only one final
checkpoint for evaluation on the hidden evaluation set. Mean is the arithmetic mean of the two
final-checkpoint scores, not best-of-two or two samples from one checkpoint. The
evaluator runs as a local process, checkpoint inference uses no-thinking mode,
and \texttt{gemini-3.5-flash} judges under an 8,192-token thinking budget and a
10\% fail-closed threshold. Checkpoint deployment uses eight GPUs and is
cleaned up after evaluation.}

\begin{table}[!ht]
\centering
\scriptsize
\caption{Final-only protocol final-checkpoint results. The terminal delta
$\Delta_{\mathrm{term}}$ compares Mean with the base score. Values are raw
evaluation scores before rollback; Mean and $\Delta_{\mathrm{term}}$ are
computed from unrounded scores before display rounding.}
\label{tab:oneshot24}
\resizebox{\linewidth}{!}{%
\begin{tabular}{@{}lrrrrrrrrrr@{}}
\toprule
& \multicolumn{5}{c}{\textbf{Qwen3.5-4B Self}} &
\multicolumn{5}{c}{\textbf{Qwen3.5-9B Self}} \\
\cmidrule(lr){2-6}\cmidrule(lr){7-11}
\textbf{Vague goal} & \textbf{Base} & \textbf{Run A} & \textbf{Run B} &
\textbf{Mean} & \textbf{$\Delta_{\mathrm{term}}$} & \textbf{Base} & \textbf{Run A} &
\textbf{Run B} & \textbf{Mean} & \textbf{$\Delta_{\mathrm{term}}$} \\
\midrule
Scientific and academic reasoning & 44.000 & 0.000 & 0.000 & 0.000 & -44.000 & 45.330 & 48.000 & 48.000 & 48.000 & \textbf{+2.670} \\
Humanities and social sciences & 18.820 & 9.000 & 5.909 & 7.455 & -11.365 & 27.450 & 10.273 & 9.091 & 9.682 & -17.768 \\
Mathematical reasoning & 17.860 & 16.746 & 0.159 & 8.452 & -9.408 & 25.160 & 0.714 & 2.540 & 1.627 & -23.533 \\
Health and medicine & 87.000 & 63.000 & 68.000 & 65.500 & -21.500 & 87.000 & 71.000 & 72.000 & 71.500 & -15.500 \\
Logic, reliability, and instruction following & 27.460 & 21.437 & 23.754 & 22.596 & -4.864 & 26.190 & 6.141 & 16.338 & 11.240 & -14.950 \\
Academic and scientific writing & 19.270 & 0.000 & 15.308 & 7.654 & -11.616 & 22.930 & 23.588 & 0.000 & 11.794 & -11.136 \\
\bottomrule
\end{tabular}}
\end{table}

\rev{All 24 submitted final checkpoints complete evaluation on the first judge attempt;
all 2,080 item outputs are nonempty and no item reports an evaluation error.
At the model--goal reporting level, Qwen3.5-4B has 0/6 Mean outcomes above its base
score and Qwen3.5-9B has 1/6. At the individual-run level, the corresponding counts are
0/12 and 3/12. Scientific and academic reasoning is the only positive Mean:
its two runs each score 48.00 and answer 36/75 items correctly, yet
only 26 correct items overlap and 20/75 items flip which run is correct. The
agreement therefore concerns aggregate direction on the same 75 evaluation items, not
item-level stability or transfer beyond those items. The raw final-checkpoint macro average
falls from 35.735 to 18.609 for 4B and from 39.010 to 25.640 for 9B; rollback
leaves the retained model equal to the base model in 21/24 runs.}

\rev{Five submitted checkpoints use numeric-label MMLU SFT. Their selected
training snapshots contain 21,000 single-digit targets, and all 279
corresponding evaluation outputs are also single digits; the five final-checkpoint
scores are 0, 0, 0, 6.141, and 0. Three long-tail runs further illustrate long
blind training lineages: 4B Run B for mathematics launches 20 jobs and spends 26.570
GPU-hours before submitting a final checkpoint that scores 0.159; 9B Run B for mathematics launches 17
jobs and spends 32.286 GPU-hours before scoring 2.540; and 9B Run B for writing launches
16 jobs and spends 30.530 GPU-hours before scoring 0. Because only final
checkpoints are evaluated, these results neither identify the best
intermediate checkpoint nor establish a monotonic degradation path.}

\subsection{Adaptive-Feedback Protocol Results}

Of the 30 configuration--goal cells in the adaptive-feedback protocol, 28/30
produce at least one evaluated checkpoint. Only two configuration--goal cells have a best
evaluated checkpoint whose score exceeds the base score: Qwen3.5-4B Self
science rises from 44.00 to 45.33, while Terra optimizing Qwen3.5-4B
mathematics rises from 17.86 to 20.10. The Terra mathematics checkpoint is the
only retained improvement under the complete-run selection rule. Among the 22 configuration--goal cells
with at least two evaluated checkpoints, 62 consecutive transitions comprise 28
increases, 13 ties, and 21 decreases. Later search beats the first evaluated
checkpoint in 14/22 configuration--goal cells, but only the Terra mathematics trajectory yields an
eligible checkpoint whose score exceeds the base score. These are descriptive cell-level results
from one run per configuration--goal cell.

\paragraph{Decision-model search profiles.}
\rev{Eight of the 22 multi-checkpoint trajectories are non-decreasing.
Luna uses SFT in 24/28 registered plans and continues from the latest checkpoint
in 21/28, producing 22 evaluated checkpoints with 89.57 settled GPU-hours. Terra
registers 33 plans, including six GRPO updates, and produces 19 evaluated
checkpoints with 91.73 GPU-hours. Its retained mathematics score of 20.10 is
followed by 2.78 and then 18.17. Sol registers 40 plans, 36 evaluation attempts,
23 search actions, and 12 distinct dataset names; it produces 33 evaluated
checkpoints with 76.56 GPU-hours, but none exceeds the corresponding base score.}

\paragraph{Self-directed proxy and update choices.}
\rev{Qwen3.5-4B Self imports GSM8K or Hendrycks mathematics data in 30/32
dataset-import events, including events for science, logic, and writing;
Qwen3.5-9B Self uses mathematics data in 11/15 imports but is more
goal-specific outside mathematics. Their update choices differ as well:
Qwen3.5-4B Self uses GRPO in 4/12 plans, whereas Qwen3.5-9B Self uses SFT in
9/11. The two configurations generate 20,396 and 28,135 agent events but
register only 12 and 11 training plans, respectively. These counts describe
search allocation and do not by themselves establish why a checkpoint gains or
loses capability.}

\subsection{Trajectory Evidence Boundary}

\rev{The portable adaptive-feedback artifact contains counts, timestamps, source hashes,
resource use, submissions, and selection records, but excludes hidden evaluation items,
raw model messages, command outputs, and per-item judge reasons. Deterministic
extraction covers every configuration--goal cell, reading 63,973 agent, 1,196 training,
and 2,194 phase events and recovering 124 training plans and 95
candidate-evaluation attempts. Two independent reviews check the task-level
interpretation. The resulting claims therefore concern exploration,
checkpoint progression, selection, and rollback; item-level causal analysis
requires controlled access to the content-bearing traces.}

\subsection{Compute and Resource Accounting}
\label{app:cost}

\begin{table}[!ht]
\centering
\small
\caption{Resource accounting for RQ2--RQ3. Training GPU-hours exclude
inference, judging, deployment, and idle controller time.}
\label{tab:compute-accounting}
\resizebox{\linewidth}{!}{%
\begin{tabular}{@{}lrrrrl@{}}
\toprule
\textbf{Experiment} & \textbf{Runs} & \textbf{Plans} & \textbf{Jobs} &
\textbf{Train GPU-h} & \textbf{Replication and cap} \\
\midrule
Adaptive-feedback protocol & 30 & 124 & 107 & 322.610 & One run/configuration--goal cell; 40 GPU-h and 10 h wall time/configuration--goal cell \\
Final-only protocol & 24 & -- & 83 & 149.236 & Separately executed Run A and Run B/model--goal cell; 40 GPU-h/run \\
RQ3 successor-harness creation & 4 creation attempts & -- & -- & 0 & 3 valid successors; three executions/successor harness \\
\bottomrule
\end{tabular}}
\end{table}

\rev{Table~\ref{tab:compute-accounting} summarizes runs, plans, jobs, and
settled training GPU-hours. The adaptive-feedback protocol is a complete $5\times6$ design but not a multi-seed
estimate. The final-only protocol uses 59.269 training GPU-hours for Qwen3.5-4B and 89.967
for Qwen3.5-9B. RQ3 does not update weights; creator-model and runtime inference
therefore lie outside the training-GPU ledger.}

\section{RQ3 Creator-Trajectory Case Analysis}
\label{app:rq3-traces}

\paragraph{Scope and evidence boundary.}
\rev{This content-level analysis covers the three valid GPT-5.6 creator
trajectories and one inspected execution on the hidden evaluation set for each frozen
successor harness. The creator sees only its own validation data; evaluation items and
scores are not returned. Table~\ref{tab:rq3} reports means over three executions,
whereas the task-level examples below diagnose mechanisms in one inspected
execution. They are therefore case evidence, not additional repetitions or a
stable ranking of creator models.}

\paragraph{Luna: early stopping on a narrow validation proxy.}
\rev{Luna evaluates $H_0$ on one validation prompt with an eight-item checklist.
After a 7/8 result, it adds a fixed five-part template covering the research
question, design, analysis, risks, and limitations. The edited harness reaches
8/8 on that prompt and two similar prompts, and Luna stops after approximately
40 minutes. In the inspected execution on the hidden evaluation set, this template improves
structural completeness but reframes direct questions about mechanisms,
formulas, parameters, or engineering details as study-design problems rather
than supplying the requested technical content.}

\paragraph{Terra: reviewer reversion without a final-answer invariant.}
\rev{Terra removes an initial two-pass reviewer after it invents unsupported
significance results and effect sizes, then adds narrower safeguards for
assumptions, model calls, logging, and calculator use. It nevertheless omits a
final-answer invariant: when the post-calculator response is empty, the harness
submits the pre-tool fragment. Two inspected tasks therefore return partial
drafts of 342 and 520 characters, with scores falling from 75.00 and 52.33
under $H_0$ to 2.57 and 2.91. These cases account for approximately 70\% of the
aggregate decline in that execution.}

\paragraph{Sol: broader testing followed by conservative selection.}
\rev{Across 3 hours 11 minutes and four validation task types, Sol tests prompt
length, tool-call recovery, disabled tools, and a review pass, while correcting
one erroneous local reference answer. It rejects locally stronger variants for
latency, repeated calls, fabricated numbers, or scorer misses. The selected
harness restores the original Qwen-Agent policy and adds a completeness check
for empty final responses. The check never triggers on the 20 inspected evaluation
items, supporting improved failure semantics rather than a score gain.}

\endgroup

\end{document}